\documentclass[letterpaper, 10 pt, conference]{IEEEtran}

\IEEEoverridecommandlockouts

\usepackage{amsmath,amsfonts,amssymb}
\usepackage{amsthm}  {
      \theoremstyle{plain}
			\newtheorem{definition}{Definition}

}

\usepackage{graphicx}
\usepackage{multirow}
\usepackage{booktabs}

\newcommand{\St}{\mathcal{S}} 
\newcommand{\A}{\mathcal{A}} 
\newcommand{\Z}{\mathcal{Z}} 
\newcommand{\T}{\mathbb{T}} 
\newcommand{\Ob}{\mathbb{O}} 
\newcommand{\B}{\mathcal{B}} 
\newcommand{\R}{\mathbb{R}} 

\DeclareMathOperator*{\argmax}{argmax}

\usepackage{url}
\usepackage[colorlinks=true,citecolor=red,pdftex]{hyperref} 

\hypersetup{
  pdfauthor={Mikko Lauri, Eero Hein\"{a}nen, Simone Frintrop},%
  pdftitle={Multi-Robot Active Information Gathering with Periodic Communication},%
  pdfsubject={Multi-Robot Active Information Gathering with Periodic Communication},%
  pdfkeywords={dec-pomdp, pomdp, policy iteration, information planning, sensor management, robotics, active perception, active information acquisition}
}

\begin{document}

\title{\LARGE \bf Multi-Robot Active Information Gathering with Periodic Communication}

\author{Mikko Lauri\authorrefmark{1}, Eero Hein\"{a}nen\authorrefmark{2}, Simone Frintrop\authorrefmark{1}
\thanks{This work was supported by the Academy of Finland decision 268152, Optimal operation of observation systems in autonomous mobile machines.}
\thanks{\authorrefmark{1} Department of Informatics, University of Hamburg, Hamburg, Germany, {\tt\small \{lauri,frintrop\}@informatik.uni-hamburg.de}.}%
\thanks{\authorrefmark{2} Laboratory of Automation and Hydraulics, Tampere University of Technology, Tampere, Finland, {\tt\small eero.heinanen@student.tut.fi}.}%
}
\maketitle

\begin{abstract}
A team of robots sharing a common goal can benefit from coordination of the activities of team members, helping the team to reach the goal more reliably or quickly.
We address the problem of coordinating the actions of a team of robots with periodic communication capability executing an information gathering task.
We cast the problem as a multi-agent optimal decision-making problem with an information theoretic objective function.
We show that appropriate techniques for solving decentralized partially observable Markov decision processes (Dec-POMDPs) are applicable in such information gathering problems.
We quantify the usefulness of coordinated information gathering through simulation studies, and demonstrate the feasibility of the method in a real-world target tracking domain.
\end{abstract}
\section{Introduction}
\label{sec:intro}
Teams of robots are projected to be applied in a wide range of information gathering tasks, ranging from locating victims in search and rescue scenarios~\cite{Balakirsky2007} to simultaneous localization and mapping~\cite{Saeedi2016}.
To benefit most from the deployment of multiple robots, the robot team should coordinate its activities.
Coordination becomes more challenging when communication between team members is limited.
In this paper, we study how the team members can coordinate their activities in an information gathering task under periodic communication.

\begin{figure}[t!]
\centering
\includegraphics[scale=1.0]{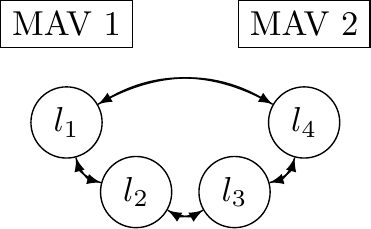}
\caption{Two micro aerial vehicles tracking a target between the locations $l_i$.}
\label{fig:example}
\end{figure}

As a motivating example, consider the scenario in Figure~\ref{fig:example} in which two robotic agents, here micro aerial vehicles (MAVs), are jointly estimating the state of a moving target.
Each MAV can activate either a vision sensor that can detect the target at close range, or a radar sensor that can detect the target when it is further away.
The MAVs can periodically communicate and share sensor data, forming a joint estimate of the target state.
Between these periods, each MAV must act without knowledge of what the other is doing.
Coordination benefits the MAVs: if the target is close to the first MAV but far from the second one, the first MAV should try to detect the target with its camera while the second MAV applies its radar sensor.
Further, the MAVs can avoid simultaneously operating their radars, avoiding  interference that may corrupt the data.

\textbf{Related work.}
Controlling robot teams in information gathering tasks is often implemented by applying various relaxations to the control problem to avoid the high computational demands.
A fully distributed algorithm applying gradient-based control with a mutual information reward is presented in~\cite{Julian2012}, and in~\cite{Zhou2011} the next best sensing locations for a robot team are found via a Gauss-Seidel relaxation.
In~\cite{Atanasov2015}, the decentralized information gathering problem is linearized while also providing suboptimality guarantees.
Periodic communication was considered through a problem constraint in~\cite{Hollinger2012}, and by designing distributed data fusion techniques that can handle communication breaks in~\cite{Hollinger2015}.
Distributed constraint optimization problems (DCOPs) have been applied to planning for coverage of multiple targets~\cite{Manish2009,Zivan2015}.
Partially observable Markov decision processes (POMDPs) were applied to change detection in~\cite{Renoux2015}, with each robot reasoning about the beliefs of the others.
Decentralized control for environmental monitoring and target tracking based on auctioning POMDP control policies is proposed in~\cite{Capitan2013}.
In~\cite{Nair2005}, DCOPs and decentralized POMDPs were exploited for control in sensor networks, where the sensor nodes only have limited local interactions. 

Common for many of the approaches above is that the underlying problem they attempt to tackle is a variant of a decentralized partially observable Markov decision process (Dec-POMDP)~\cite{Oliehoek2016}.
A Dec-POMDP is a generic model for cooperative multi-agent sequential decision making under uncertainty, where agents with a shared goal execute actions and perceive observations that provide partial information about the underlying hidden state of the system.
Dec-POMDPs explicitly model uncertainty in sensor data and system dynamics, making them an appealing model for robotics problems.
A solution to a Dec-POMDP is a control policy, computed centrally and distributed to the agents for execution.
Although finding an optimal policy for a Dec-POMDP is NEXP-hard~\cite{Bernstein2002}, recent work has been able to improve tractability at multi-robot control problems such as package delivery~\cite{Omidshafiei2015,Omidshafiei2016}.

Single-robot information gathering has also been addressed as a POMDP or a stochastic control problem, see~\cite{Atanasov2014,Charrow2014,Indelman2015,Lauri2016}.
In these approaches, information gathering is explicitly addressed by defining an information-theoretic reward function, e.g., mutual information or entropy.
Our aim here is to translate this into a multi-robot setting through Dec-POMDPs with information-theoretic rewards.

\textbf{Contribution.}
Our contribution is summarized as follows.
First, we introduce the $\rho$Dec-POMDP model, which extends the Dec-POMDP to allow information-theoretic rewards.
We show how existing Dec-POMDP solution methods can be extended to $\rho$Dec-POMDPs.
Secondly, we quantify through experiments in target tracking domains the usefulness and feasibility of our approach to decentralized information gathering.

We assume that the robots can periodically communicate to share information.
In this sense, our approach is less general compared to e.g.~\cite{Hollinger2012} who require periodic communication via constraints on the problem.
In contrast to e.g.~\cite{Nair2005,Capitan2013,Renoux2015}, we do not assume any state transition or observation independence properties. 
Compared to existing approaches for multi-robot control via Dec-POMDPs, ours differs in that we explicitly minimize a measure of state estimate uncertainty.

\textbf{Organization.}
Section~\ref{sec:decpomdp} reviews the Dec-POMDP.
In Section~\ref{sec:infodecpomdp}, we introduce our extension, the $\rho$Dec-POMDP, and discuss its properties as applied to multi-robot information gathering.
Section~\ref{sec:planning} explains how solution algorithms for standard Dec-POMDPs may be applied to $\rho$Dec-POMDPs.
Sections~\ref{sec:simulation} and~\ref{sec:experiment} report results of simulation and real-world target tracking experiments.
Section~\ref{sec:conclusion} concludes the paper.
\section{Decentralized cooperative decision making}
\label{sec:decpomdp}
We study sequential decision-making by a team of robotic agents, formalized as a decentralized partially observable Markov decision process (Dec-POMDP).
The definitions and concepts referred to in this section are based on~\cite{Oliehoek2008} and~\cite{Oliehoek2016}.
\begin{definition}[Dec-POMDP]
\label{def:decpomdp}
A Dec-POMDP is a tuple $\langle I, \St, \lbrace \A_i \rbrace_{i\in I}, \lbrace \Z_i \rbrace_{i\in I}, \T, \Ob, R, b_0 \rangle$, where
\begin{itemize}
\item $I=\{1,2,\ldots,n \}$ is the set of $n$ agents,
\item $\St$ is the finite state space,
\item $\A_i$ and $\Z_i$ are the finite action and observation spaces of agent $i$ such that $\A = \times_{i\in I} \A_i$ and $\times_{i\in I} \Z_i$ are the \emph{joint} action and observation spaces, respectively,
\item $\T\colon\St\times\A\times\St\to [0,1]$ is a stochastic state transition model, such that $\T(s',a,s)$ gives the conditional probability of transitioning to state $s'\in\St$ when joint action $a\in\A$ is executed at state $s\in\St$,
\item $\Ob\colon\Z\times\St\times\A\to[0,1]$ is a probabilistic observation model, such that $\Ob(z',s',a)$ gives the conditional probability of perceiving joint observation $z'\in\Z$ in state $s'\in\St$ when the previous joint action was $a\in\A$,
\item  $R\colon\St\times\A\to\R$ is the reward function, and
\item $b_0$ is the initial joint belief state, encoding information about the state at time $t=0$.
\end{itemize}
\end{definition}

At each time step $t$, each agent $i$ selects an action $a_t^i \in \A_i$, forming a joint action $a_t = (a_t^1, \ldots, a_t^n)$.
The agents obtain a shared reward according to $R$.
The state then transitions according to $\T$, and each agent perceives an observation $z_{t+1}^i \in \Z_i$ such that the joint observation $z_{t+1} = (z_{t+1}^1, \ldots, z_{t+1}^n)$ is distributed as specified by $\Ob$.
The objective of the agents is to select actions such that the expected sum of rewards over a given horizon of time $h>0$ is maximized.
To characterize the solution, we give some supporting definitions.
\begin{definition}[Local and joint history]
The local history of agent $i$ at time $t$ is $\theta^i_t = (a_0^i, z_1^i, \ldots, a_{t-1}^i, z_t^i  ) \in \Theta_t^i = \times_{t}(\A_i\times\Z_i)$. The joint history $\theta_t\in \Theta_t = \times_i \Theta_t^i$ is the collection of each agent's local histories: $\theta_t = (\theta_t^1, \ldots, \theta_t^n )$.
\end{definition}
\begin{definition}[Local and joint decision rule]
A local decision rule $\delta_t^i$ of agent $i$ at time $t$ is a function from its local histories to its individual actions: $\delta_t^i\colon \Theta_t^i \to \A_i$.
A joint decision rule $\delta_t$ is a collection of local decision rules: $\delta_t = (\delta_t^1, \ldots, \delta_t^n)$.
\end{definition}
\begin{definition}[Joint policy]
A joint policy $\pi \in \Pi_h$ for horizon $h$ is a collection of joint decision rules: $\pi = (\delta_0, \delta_1, \ldots, \delta_{h-1})$.
\end{definition}
Thus, $\pi$ is a mapping $\Theta_t \to \A$ for any $t$ s.t.\ $0 \leq t < h$.
The value $V(\pi)$ of a joint policy is given by
\begin{equation}
\label{eq:decpolicy_value}
V(\pi) = \sum\limits_{t=0}^{h-1}\sum\limits_{\theta_t \in \Theta_t} P\left(\theta_t \,\middle|\, \pi, b_0 \right) R(\theta_t, \pi(\theta_t)).
\end{equation}
Here, $P\left(\theta_t \,\middle|\, \pi, b_0 \right)$ is the probability of experiencing joint history $\theta_t$ when executing policy $\pi$ starting from $b_0$.
Furthermore, $R(\theta_t, a_t) = \sum\limits_{s_t\in \St} R(s_t, a_t)P\left(s_t \,\middle|\, \theta_t, b_0\right)$ is the expected immediate reward of executing joint action $a_t$ after experiencing joint history $\theta_t$.
An optimal solution is a joint policy $\pi^*$ with a maximal value: $\pi^* = \argmax\limits_{\pi \in \Pi_h} V(\pi)$.

The probability mass function (pmf) $P\left(\cdot \,\middle|\, \theta_t, b_0\right)$ is a sufficient statistic for the state given the joint history $\theta_t=(a_0, z_1, \ldots, a_{t-1}, z_t)$ and the joint belief state $b_0$.
We denote  $P\left(\cdot \,\middle|\, \theta_t, b_0\right) \equiv b_t$, and by $\B$ the space of all such pmfs over the state space $\St$.
Joint belief states are computed recursively applying the Bayesian filtering operator $\tau\colon\B\times\A\times\Z\to\B$:
\begin{equation}
\label{eq:belief_update}
\begin{split}
b_t& = \tau(b_{t-1},a_{t-1},z_{t}) \equiv \frac{1}{\eta(z_{t}\mid b_{t-1}, a_{t-1})} \cdot \\
& \Ob(z_t,s_{t},a_{t-1})\cdot\sum\limits_{s_{t-1}\in\St}\T(s_t,a_{t-1},s_{t-1})b_{t-1}(s_{t-1}),
\end{split}
\end{equation}
where $\eta(z_t\mid b_{t-1}, a_{t-1})$ is the normalizing factor equal to the prior probability of perceiving $z_{t}$ when joint action $a_{t-1}$ is taken in joint belief state $b_{t-1}$.
For later use, define
\begin{equation}
\label{eq:belief_update_history}
b_t = \tau(\theta_t, b_0) \equiv \tau( \tau( \ldots , a_{t-2}, z_{t-1}), a_{t-1}, z_t)
\end{equation}
as an equivalent shorthand notation for expressing the joint belief state $b_t$ as function of the joint history $\theta_t$ and $b_0$.
\section{Decentralized information gathering}
\label{sec:infodecpomdp}
We next present our extension of the Dec-POMDP model for information-theoretic rewards.
Additionally, we motivate and discuss our assumption of periodic communication.

\subsection{Information-theoretic rewards for Dec-POMDPs}
Consider an information gathering task for a robot team.
It seems clear that an optimal policy should lead the robots to act such as to reach a state estimate with low uncertainty.
Reward functions that only depend on the state and action do not appear to be fully compatible with such objectives~\cite{Araya2010}.
For example, consider a robot team collecting information about a physical process which they are unable to affect through their actions, such as monitoring underwater ocean currents.
In this case, the underlying state is not meaningful for the robots' objective, nor are the possible actions by themselves.

In single-agent POMDPs, information gathering has been addressed, e.g., by augmenting the action space to include new actions that reward information-gathering~\cite{Spaan2010,Spaan2015}, or by applying information-theoretic rewards~\cite{Araya2010}.
The first approach results in a multiplication of the size of the agents' action spaces $\A_i$.
As the number of possible policies for an agent $i$ in a Dec-POMDP is $|\A_i|^{((|\A_i||\Z_i|)^h-1) / (|\A_i||\Z_i|-1))}$~\cite{Oliehoek2008}, this approach does not seem promising to translate to Dec-POMDPs.
We consider the second approach, which corresponds to setting a reward function that depends on the joint belief state instead of the true underlying state of the system.

In a Dec-POMDP, an individual agent usually cannot compute a joint belief state  based only on its local history.
However, during the centralized planning phase the possible joint histories $\theta_t$ are available, and each joint history corresponds to some joint belief state as indicated by Eq.~\eqref{eq:belief_update_history}.
Thus, information theoretic reward functions may be evaluated and applied in Dec-POMDPs while computing a joint policy.
Inspired by~\cite{Araya2010}, we call the resulting model a $\rho$Dec-POMDP.
\begin{definition}[$\rho$Dec-POMDP]
\label{def:rhodecpomdp}
A $\rho$Dec-POMDP is a tuple $\langle I$, $\St$, $\lbrace \A_i \rbrace_{i\in I}$, $\lbrace \Z_i \rbrace_{i\in I}$, $\T$, $\Ob$, $\rho$, $b_0 \rangle$, where $I$, $\St$, $\lbrace \A_i \rbrace_{i\in I}$, $\lbrace \Z_i \rbrace_{i\in I}$, $\T$, $\Ob$, and $b_0$ are as in the Dec-POMDP, and the reward function is $\rho:\B\times\A\to\R$.
\end{definition}
Choosing $\rho(b,a) = \sum\limits_{s\in\St}R(s,a)b(s)$, the definition above subsumes the standard Dec-POMDP (Definition~\ref{def:decpomdp}).
As in Eq.~\eqref{eq:decpolicy_value}, the value of a joint policy $\pi$ in a $\rho$Dec-POMDP is 
\begin{equation}
\label{eq:rhodecpolicy_value}
V(\pi) = \sum\limits_{t=0}^{h-1}\sum\limits_{\theta_t \in \Theta_t} P\left(\theta_t \,\middle|\, \pi, b_0 \right) \rho( \tau(\theta_t, b_0), \pi(\theta_t)),
\end{equation}
where $\tau(\theta_t, b_0)$ is the joint belief state computed by Eq.~\eqref{eq:belief_update_history}.

An appropriate reward function encourages the agents to reach joint histories corresponding to joint belief states with low uncertainty.
This is quantified via uncertainty functions.
\begin{definition}[Uncertainty function~\cite{DeGroot2004}]
Any non-negative, concave function $g:\B\to\R^+$ is applicable as an uncertainty function.
\end{definition}
An example of an uncertainty function is the Shannon entropy $H(b) = -\sum\limits_{s}b(s) \log_2 b(s)$~\cite{Cover2006}.
An uncertainty function has a small value for joint belief states near the degenerate case where all probability mass is concentrated on one state, and a greater value near the uniform distribution.

Throughout the remainder of the paper, we will consider $\rho$Dec-POMDP models with reward functions defined as
\begin{equation}
\label{eq:rhoreward}
\rho(b,a) = \sum\limits_{s\in\St}R(s,a)b(s) -\alpha g(b),
\end{equation}
where $R(s,a)$ models the rewards dependent on states and actions only, and $g$ is an uncertainty function encoding the information gathering objective\footnote{We apply a minus sign here to penalize for uncertainty in $b$.}.
The term $\alpha > 0$ sets the balance of state-dependent and information gathering rewards.

\subsection{Periodic communication}
\begin{figure}[t]
\centering
\includegraphics[width=0.8\columnwidth]{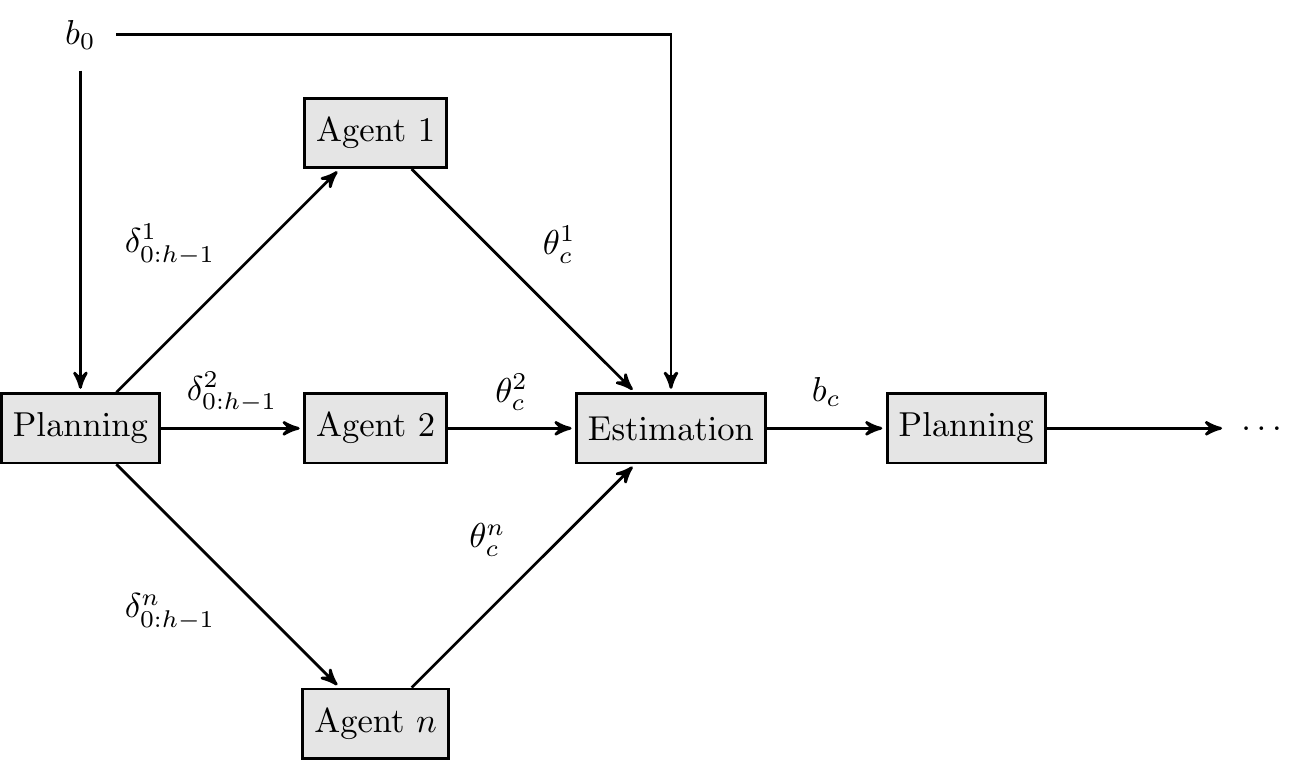}
\caption{Decentralized information gathering with communication period $c$. The agents act according to local decision rules $\delta_{0:h-1}^i$ for $c$ time steps ($c \leq h$), and then share their local histories $\theta_{c}^i$.
A joint belief state $b_{c}$ may then be estimated, and the information applied to plan subsequent actions.}
\label{fig:decpolicy}
\end{figure}

Deploying a team of robots to collect information is useful if the collected information will eventually be combined to form a joint state estimate.
We assume the robots share data via periodic communication every $c$ time steps.
This leads to a scheme outlined in Figure~\ref{fig:decpolicy}.
Given an initial joint belief state, a joint policy is computed for a horizon $h$, and distributed to the agents for execution.
After $c$ time steps, the agents communicate to share their local histories, enabling estimation of a new joint belief state.
The new information may be applied to revise the joint policy.
The communication period $c$ need not be equal to the horizon $h$.

For $c=1$, information is immediately shared by all agents and the problem reduces to a multi-agent POMDP~\cite{Pynadath2002}.
The case $c>1$ is similar to the $k$-steps delayed communication case~\cite{Oliehoek2016}: at time $t$ all agents know $\theta_{t-k}$.
However, for periodic communication this information is obtained only during the communication intervals, instead of at every time step.

One might argue that it is sufficient to set a non-zero information reward $g$ only for the last time instant of the planning horizon.
While appropriate in some applications, a counterargument can be presented for applications such as the robot team monitoring ocean currents.
Here, we wish to estimate the state of the process accurately at all time steps to learn about the process dynamics.
In such cases, minimizing the ``average uncertainty'' via Eq.~\eqref{eq:rhoreward} is reasonable.
\section{Planning in $\rho$Dec-POMDPs}
\label{sec:planning}
A planning algorithm for standard Dec-POMDPs can be applied to $\rho$Dec-POMDPs if it does not rest on the assumption that the reward is state and action dependent only.
For instance, approaches that cast the problem as a search through the joint policy space $\Pi_h$ can be applied once reward function evaluation is modified appropriately.
For example, variants of generalized multi-agent A* algorithms~\cite{Oliehoek2008} are applicable.

We applied the multi-agent A* (MAA*) algorithm~\cite{Szer2005}.
MAA* is a heuristic search algorithm similar to the classic A* algorithm.
Starting with $t=1$, MAA* constructs a search tree over partial joint policies $\phi_t = (\delta_0, \delta_1, \ldots, \delta_{t-1})$ with $t < h$ that specify how the agents act until time step $t$.
Thus, each node in the search tree represents a partial joint policy.
The search tree is expanded to cover the partial joint policies $\phi_{t+1}$ by appending to $\phi_t$ any possible joint decision rule: $\phi_{t+1} = (\phi_t, \delta_t)$.
The tree is expanded in a best-first order, as determined by a value estimate of the nodes currently in the tree.
The value estimate $\hat{V}(\phi_t)$ of a node with partial joint policy $\phi_t$ is computed in two parts: by exact evaluation of $\phi_t$ plus an estimate of the value of the remaining $h-t$ time steps via a heuristic function $H_{h-t}$: $\hat{V}(\phi_t) = V(\phi_t) + H_{h-t}(\phi_t)$.
Here, $V(\phi_t)$ is the value of $\phi_t$ computed via Eq.~\eqref{eq:rhodecpolicy_value}, and $H_{h-t}(\phi_t)$ is the heuristic value. 
If $H_{h-t}$ overestimates the true expected reward over the remaining $h-t$ time steps, MAA* returns an optimal policy~\cite{Szer2005}.

To define a heuristic function, the Dec-POMDP is relaxed, e.g., into a centralized POMDP, or into a fully observable centralized Markov decision process (MDP)~\cite{Oliehoek2008}.
The heuristic function is obtained, e.g., in the case of POMDP relaxation, by finding the optimal value of the POMDP.
We apply heuristic functions obtained via a POMDP relaxation.
This preserves the uncertainty aspect in the problem, and the optimal value can be computed even for information-theoretic reward functions by existing techniques~\cite{Araya2010}.
\section{Simulation experiments}
\label{sec:simulation}
In this section, we study the usefulness of coordinating information gathering activities in a target tracking domain.
To the best of our knowledge, there are currently no methods directly comparable to ours that address the same problem.
The related approaches~\cite{Nair2005,Renoux2015} require transition and observation independence, and the auctioning method proposed in~\cite{Capitan2013} and the multi-robot Dec-POMDP studies~\cite{Omidshafiei2015,Omidshafiei2016} use state and action based rewards.
Instead, we compare optimal $\rho$Dec-POMDP policies to hand-tuned heuristic policies.

\subsection{Cooperative target tracking domain} 
In the scenario of Figure~\ref{fig:example}, the state consists of the target location $l \in \{l_1, l_2, l_3, l_4\} = L$ and a binary variable describing the target status; neutral (0) or hostile (1).
The target does not change its status, but moves in a different pattern depending on the status.
A neutral target stays in place with probability $p_0$, and moves to either neighbouring location with probability $(1-p_0)/2$.
A hostile target stays in place with probability $p_1$.
We set $p_0 = 0.85$ and $p_1 = 0.6$.

Both MAVs have two actions, $a_c$ and $a_r$, referring to applying a camera or a radar, respectively.
The observations $\Z_1=\Z_2=L$ correspond to perceiving the target at any of the locations.
The observations do not provide information on the target status, which has to be inferred based on a series of observations.
Sensor accuracy depends on the distance to the target.
At $l_1$, the target is at distance 0 from MAV 1, and at distance 3 from MAV 2, at $l_2$, the distances are 1 and 2, and so on.
We model the sensors by Gaussian distributions with mean at the target location, and standard deviation dependent on the target status and increasing as a function of the distance.
For a sensor mode $j$, given a distance $d$, its standard deviation is $\sigma_j(d) = \sigma_{j,0} \cdot 2^{(d/d_{j,0})}$, where $\sigma_{j,0}$ is the nominal standard deviation, and $d_{j,0}$ is the half efficiency distance of the sensor.
The parameters we applied are summarized in Table~\ref{tab:obsmodel}.

\begin{table}[]
\centering
\caption{Sensor parameters in the simulation.}
\label{tab:obsmodel}
\begin{tabular}{@{}ccccc@{}}
\toprule
            & \multicolumn{2}{c}{Neutral target} & \multicolumn{2}{c}{Hostile target} \\ \midrule
Sensor mode $j$ & $d_{j,0}$      & $\sigma_{j,0}$  & $d_{j,0}$       & $\sigma_{j,0}$            \\ \midrule
camera              &   0.6      &  0.3            &   0.7           & 0.75             \\
radar               &  1.0       &  0.2            &   1.0           & 0.45             \\
radar (interference)&  2.0       &  1.0            &   1.5           & 1.2             \\ \bottomrule
\end{tabular}
\end{table}

The reward is as in Eq.~\eqref{eq:rhoreward}, with $\alpha = 1$ and $g$ as Shannon entropy.
In $R(s,a)$, for each agent applying the radar action $a_r$, there is a reward of -0.1, and additionally, if the target is hostile, an additional reward of -1 or -0.1 if the target is at distance 0 or 1, respectively.
This term models the higher costs of applying the radar, and the risk of revealing the MAVs' own location to a hostile target if radar is engaged at short range.

\subsection{Experimental results}
We varied the communication interval $c$ and the initial joint belief state $b_0$.
We compare the value of an optimal policy of the $\rho$Dec-POMDP\footnote{We apply MAA* from the MADP toolbox \url{www.fransoliehoek.net/madp/}, extended by us to handle information theoretic rewards.} to five heuristic policies.
Policy 1 is a risk-averse strategy where both MAVs only apply cameras.
In policy 2, the first MAV always applies its camera, and the second MAV always applies its radar.
Policy 3 is the same as policy 2, reversing the roles.
Policy 4 is a turn-taking policy where the first MAV starts by applying its camera while the second MAV applies its radar, and on subsequent time steps they switch sensors; policy 5 is the same with reversed roles.

\begin{figure}[t]
\centering
\includegraphics[width=0.8\columnwidth]{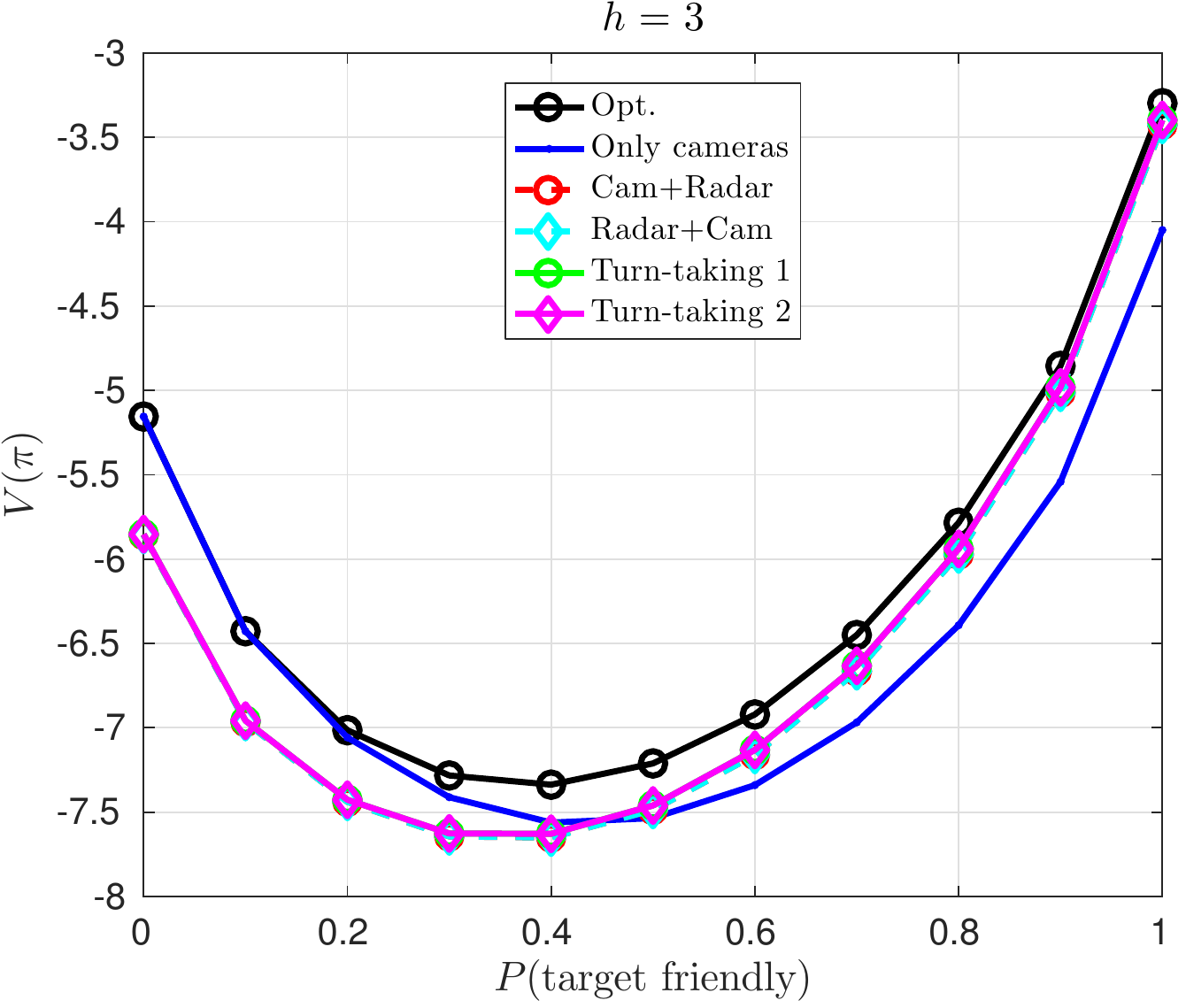}
\caption{Values of optimal and heuristic policies as a function of the initial probability of the target being neutral. The policies corresponding to the red, cyan, green and magenta lines have almost equal values and overlap each other.}
\label{fig:pfriendly}
\end{figure}

We set $b_0$ uniform with respect to the target's location, and varied the initial probability that the target is neutral.
Figure~\ref{fig:pfriendly} shows the values of the policies for $h=3$ as function of the probability that the target is neutral.
If it is very likely that the target is hostile, the heuristic policy of only applying the cameras is close to optimal as it avoids the risk of additional costs for radar use on hostile targets.
The cameras only policy has a much lower value when it is more likely that the target is neutral.
In this case, the other heuristic policies all work equally well, and are close to optimal.
None of the heuristic policies can consistently reach near-optimal performance.
An important advantage of an optimal policy is that it adapts to changing situations, unlike the fixed heuristic policies.

We then set $b_0$ uniform, and varied the communication interval $c$ between 1 and 3.
At every $c$th time step, a new policy for horizon $h=3$ was computed and then applied for the next $c$ decisions.
We ran 50 simulations on the task, each for 51 decisions, recording the rewards for each policy.
Table~\ref{tab:rewards} shows the average total rewards with 95\% confidence intervals for an optimal policy for $h=3$ while varying $c$, each heuristic policy, and a policy of choosing random actions.
Policies 2 and 3 are grouped together as ``fixed roles'', and 4 and 5 as ``turn-taking'' as there was no significant difference between them.
Here the optimal policy performs equally well regardless of the communication interval, while the turn-taking policy also performs well.
The lack of improvement for lower $c$ is explained by the fact that the actual horizon of the task is equal to 51, the number of decisions to be taken, which is much longer than the planning horizon applied.

\begin{table}[]
\centering
\caption{Average rewards with 95\% confidence intervals.}
\label{tab:rewards}
\begin{tabular}{@{}ccc@{}}
\toprule
Policy             & Comm. interval & Reward \\ \midrule
\multirow{3}{*}{Optimal} & 1  &  -89.9 $\pm$ 1.2    \\
                   & 2  & -90.1 $\pm$ 1.6   \\
                   & 3  &  -89.8 $\pm$ 1.2    \\ \midrule
Cameras only        & - &  -96.6 $\pm$ 2.0    \\
Fixed roles        & - &  -95.0 $\pm$ 1.5   \\
Turn-taking        & - &  -90.7 $\pm$ 1.4   \\
Random             & - &  -104.2 $\pm$ 2.1   \\ \bottomrule
\end{tabular}
\end{table}

\section{Cooperative tracking}
\label{sec:experiment}
\begin{figure}[t]
\centering
\includegraphics[width=0.8\columnwidth]{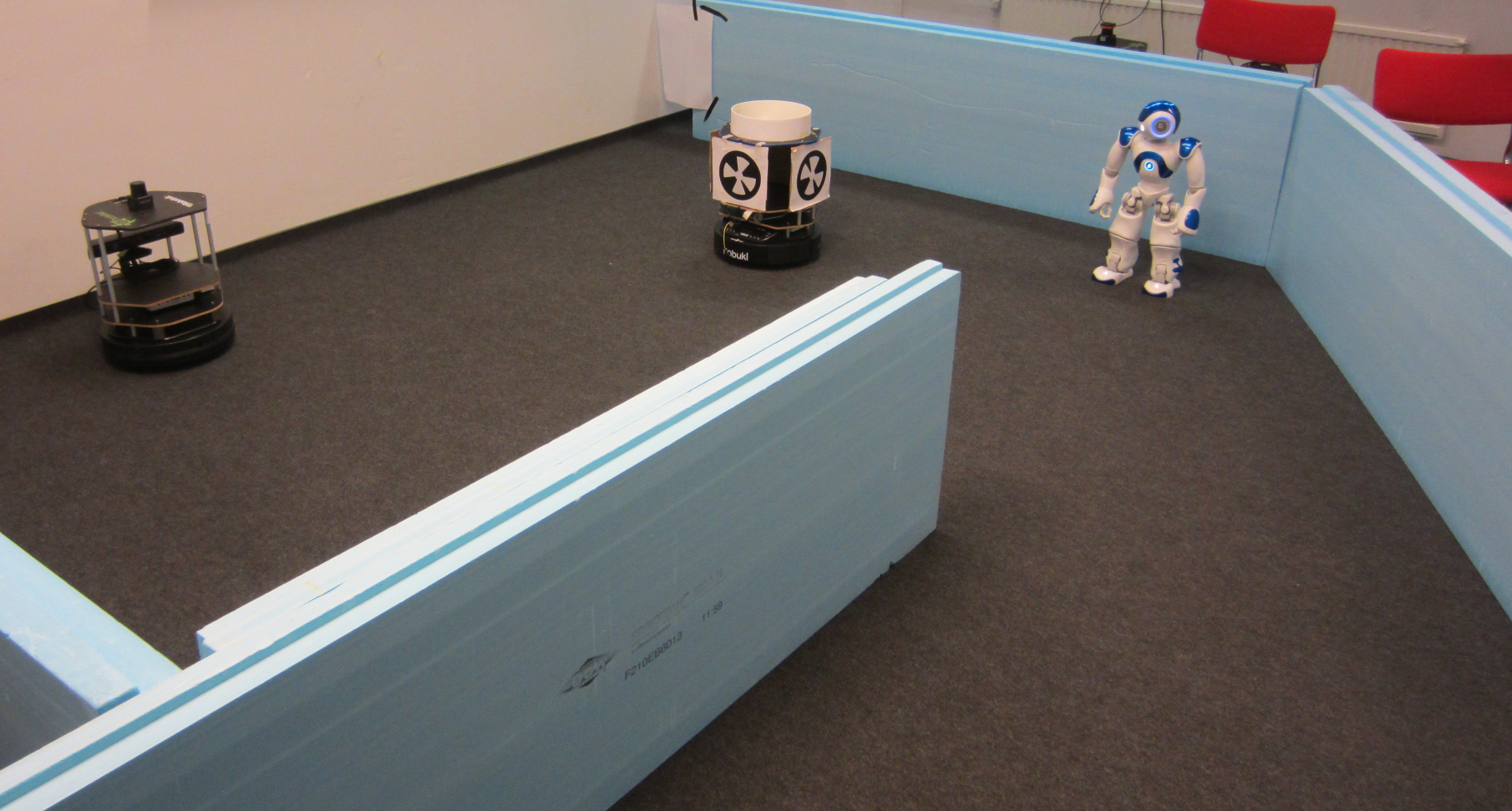}
\caption{The robots on the left and right track the target robot at the center.}
\label{fig:experiment}
\end{figure}

We set up a target tracking experiment as shown in Figure~\ref{fig:experiment}.
The target in the center of the figure was programmed to move randomly in the area.
The markers on the target can be detected by the robot on the left applying its laser range finder, and by the robot on the right applying its camera.

\begin{figure}[t]
\centering
\includegraphics[width=0.8\columnwidth]{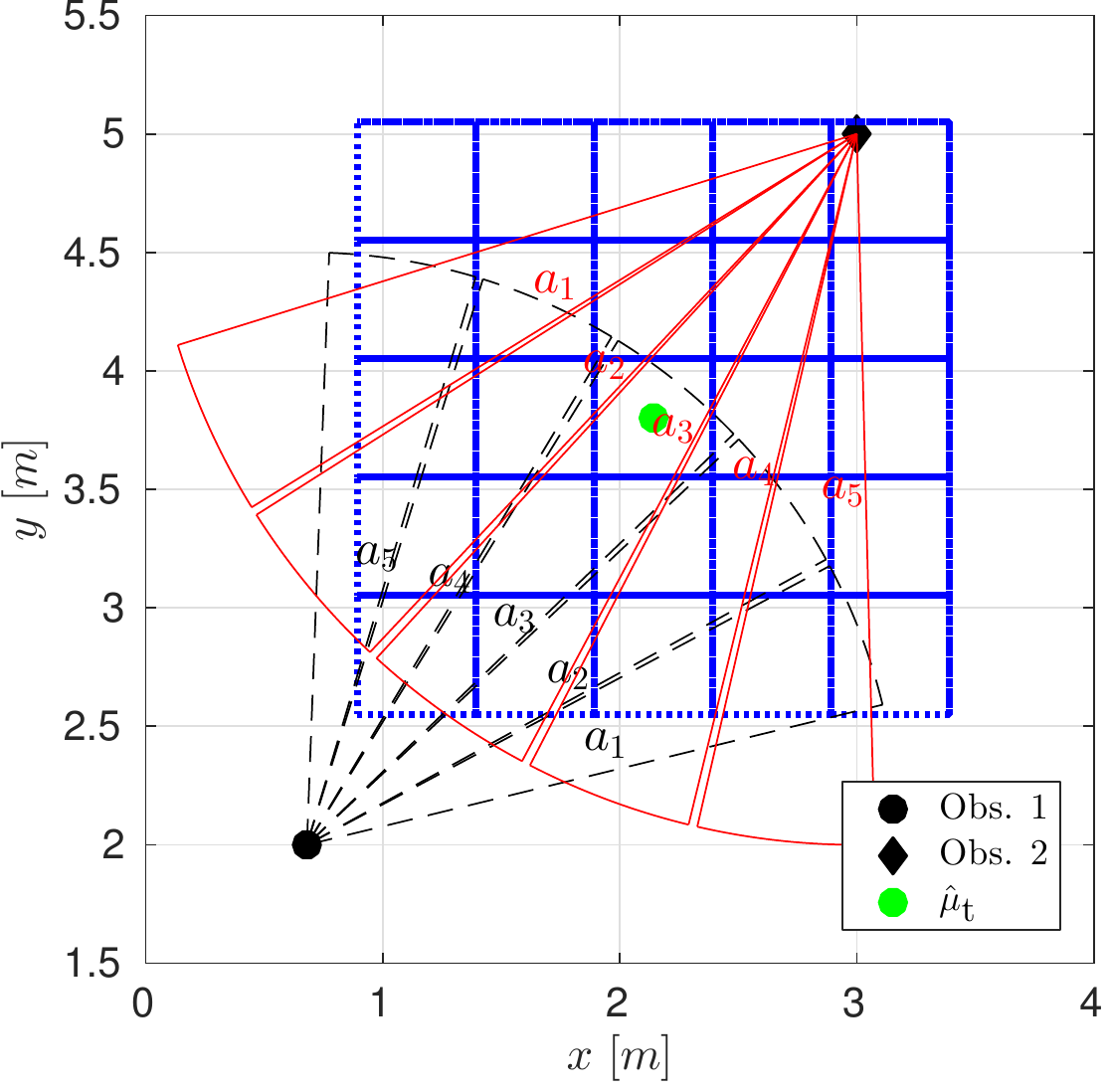}
\caption{One of five detection sectors $a_1$ through $a_5$ is chosen by each of the two observers. The Kalman filter position estimate is mapped to a finite grid, shown in blue, centred at the estimate mean $\hat{\mu}_{\textnormal{t}}$.}
\label{fig:experimental_illustration}
\end{figure}

To study cooperative target tracking, we applied a Kalman filter (KF) to estimate the target's position and velocity while limiting the amount of input data.
A schematic of the experimental setup is shown in Figure~\ref{fig:experimental_illustration}.
At each 1-second time interval, both observers select a detection sector to focus their attention on, as indicated by the labels $a_1$ through $a_5$.
We only input an observation of the target to the KF if it was within the selected detection sector.
Each detection sector was 15 degrees wide, with a maximum range of 2.5 and 3 meters, respectively.
Selecting overlapping detection sectors could result in interference corrupting the data.

As the KF state estimate is continuous, at each time instant we mapped the tracking task into a finite $\rho$Dec-POMDP.
The KF state estimate was discretized to a 5-by-5 grid centred at the mean $\hat{\mu}_t$ of the current target position estimate.
The grid cell size was adaptively tuned so the grid covered the 3-sigma range of the estimate.
The detection sectors were defined always setting the middle sector's center to point towards $\hat{\mu}_t$.
For target motion we assumed a Gaussian velocity distribution, with mean equal to the velocity estimate from the KF, and covariance matching the expected velocity of the target.
The observations indicated if the target was detected or not within the selected detection sector.
There was a nominal false negative probability of 0.15, and a false positive probability of 0.05.
If overlapping detection sectors were selected, these probabilities were increased in proportion to the area of overlap, up to a maximum value of 0.5.
The reward was as in Eq.~\eqref{eq:rhoreward}, $R$ all zero, $\alpha=1$, and $g$ as Shannon entropy.

The $\rho$Dec-POMDP optimal policy was computed with $h=3$ and applied to select detection sectors over the next $c=3$ time steps.
We compared this to a policy where the first robot repeated $a_1$, $a_2$, $\ldots$, $a_5$, while the second robot repeated this sequence in reversed order, and a random policy.
We modelled interference due to overlapping detection sectors by corrupting the observations according to a probability proportional to the area of overlap.
A baseline was computed applying all data in a KF, without regard to the detection sectors.

\begin{figure}[t]
\centering
\includegraphics[width=0.8\columnwidth]{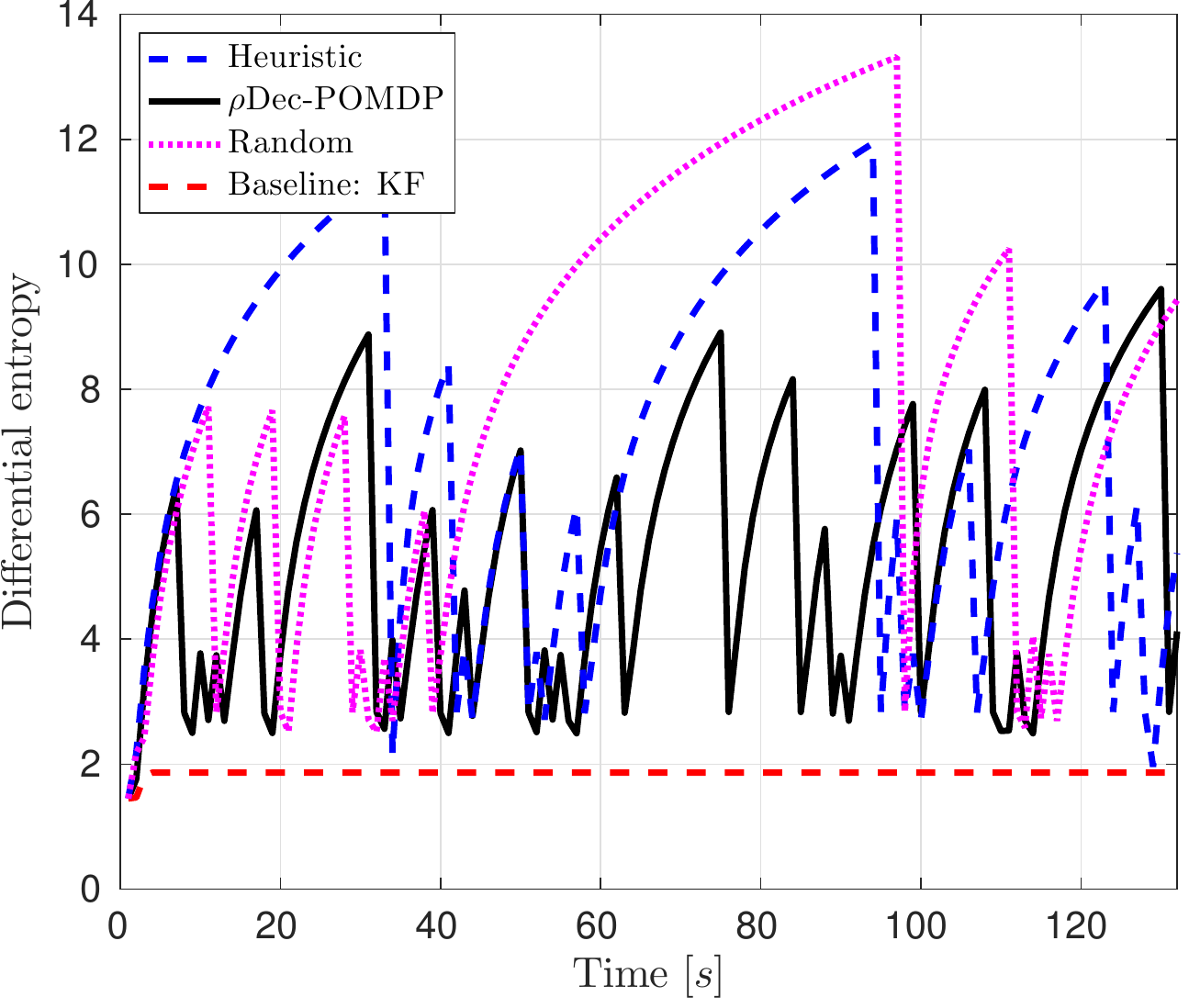}
\caption{Differential entropy of the position estimate (lower is less uncertain) as function of time.}
\label{fig:tracking_entropy}
\end{figure}

Figure~\ref{fig:tracking_entropy} shows the differential entropy~\cite{Cover2006} of the position estimate as function of time.
Due to its adaptivity, the $\rho$Dec-POMDP method performs better than the heuristic or random policy, maintaining a lower average entropy.
Interference occurred during 2 time steps for $\rho$Dec-POMDP, and during 41 time steps for the heuristic policy, and 5 time steps for the random policy.
Thus, the $\rho$Dec-POMDP policy avoids the risk of corrupting observations due to selecting overlapping detection sectors.
Compared to the baseline KF estimate, the $\rho$Dec-POMDP policy had a sum of squared position error of 143.4, the heuristic policy 178.4, and the random policy 519.7.

\section{Conclusion}
\label{sec:conclusion}
For modelling information gathering by a robot team, we presented $\rho$Dec-POMDP, extending the Dec-POMDP to information-theoretic rewards.
A $\rho$Dec-POMDPs may be solved applying existing algorithms for Dec-POMDPs, with modified reward function evaluation.
We verified the feasibility of our approach for cooperative target tracking.
Due to the adaptivity of $\rho$Dec-POMDP policies, they can outperform heuristic approaches.
Future work includes extended empirical evaluation, and possibly combining $\rho$Dec-POMDPs with distributed state estimation to relax communication assumptions.

\bibliographystyle{IEEEtran}
\bibliography{ref}

\end{document}